# Automated GIS-Based Framework for Detecting Crosswalk Changes from Bi-Temporal High-Resolution Aerial Images

**Richard Boadu Antwi**
Graduate Research Assistant
Department of Civil and Environmental Engineering, Florida A&M University–Florida State University
College of Engineering, Florida State University
2525 Pottsdamer Street, Tallahassee, FL 32310, USA
Corresponding author, Email: rantwi@fsu.edu

**Samuel Takyi**
Graduate Research Assistant
Department of Civil and Environmental Engineering, Florida A&M University–Florida State University
College of Engineering, Florida State University
2525 Pottsdamer Street, Tallahassee, FL 32310, USA

**Alican Karaer**
Associate Data Scientist
Iteris, Inc.
Tallahassee, FL, 32304, USA

**Eren Erman Ozguven**
Associate Professor
Department of Civil and Environmental Engineering, Florida A&M University–Florida State University
College of Engineering, Florida State University
2525 Pottsdamer Street, Tallahassee, FL 32310, USA

**Michael Kimollo**
Graduate Student
School of Engineering, University of North Florida
Jacksonville, FL 32224

**Ren Moses**
Professor
Department of Civil and Environmental Engineering, Florida A&M University–Florida State University
College of Engineering, Florida State University
2525 Pottsdamer Street, Tallahassee, FL 32310, USA

**Maxim A. Dulebenets**
Associate Professor
Department of Civil and Environmental Engineering, Florida A&M University–Florida State University
College of Engineering, Florida State University
2525 Pottsdamer Street, Tallahassee, FL 32310, USA

**Thobias Sando**
Professor
School of Engineering, University of North Florida
Jacksonville, FL 32224

**ABSTRACT**

Identification of changes in pavement markings has become crucial for infrastructure monitoring, maintenance, development, traffic management, and safety. Automated extraction of roadway geometry is critical in helping with this, given the increasing availability of high-resolution images and advancements in computer vision and object detection. Specifically, due to the substantial volume of satellite and high-resolution aerial images captured at different time instances, change detection has become a viable solution. In this study, an automated framework is developed to detect changes in crosswalks of Orange, Osceola, and Seminole counties in Florida, utilizing data extracted from high-resolution images obtained at various time intervals. Specifically, for Orange County, crosswalk changes between 2019 and 2021 were manually extracted, verified, and categorized as either new or modified crosswalks. For Seminole County, the developed model was used to automatically extract crosswalk changes between 2018 and 2021, while for Osceola County, changes between 2019 and 2020 were extracted. Findings indicate that Orange County witnessed approximately 2,094 crosswalk changes, with 312 occurring on state roads. In Seminole and Osceola counties, on the other hand, 1,040 and 1,402 crosswalk changes were observed on both local and state roads, respectively. Among these, 340 and 344 were identified on state roads in Seminole and Osceola, respectively. Spatiotemporal changes observed in crosswalks can be utilized to regularly update the existing crosswalk inventories, which is essential for agencies engaged in traffic and safety studies. Data extracted from these crosswalk changes can be combined with traffic and crash data to provide valuable insights to policymakers.

**INTRODUCTION**

The field of computer vision technology is making rapid progress, enabling traffic agencies to achieve cost and time savings in collecting roadway geometry data. According to a study conducted by (*1-3*), aerial and satellite imagery proved to be more favorable than field observations concerning equipment cost, data accuracy, crew safety, and data collection cost and time. In the past, image processing was acknowledged as a time-consuming and error-prone approach for acquiring roadway data. However, recent advancements in computing power and image recognition methods have introduced exciting prospects for accurately detecting and mapping various roadway features from diverse imagery data. Remote sensing research focused on change detection (CD), image classification and analysis are significant undertakings (*4-6*). Change detection is employed to detect changes or modifications occurring within a geographic area by combining multiple images captured at different points in time (*9*). The objective of a change detection algorithm is to identify significant changes while disregarding unimportant ones (*8*). The increased accessibility of extensive collections of historical images enables the feasibility of long-term change detection and modelling. This has stimulated the development in change detection research following the availability of satellite and high-resolution aerial images of a particular geographical area taken at different time instances. This advancement encourages additional exploration into the development of advanced image processing techniques and development of novel approaches for effectively managing image data over time (*9*). Change detection studies are motivated by various significant factors, which can be attributed to shifts in environmental conditions, meteorological patterns, and physical development (*10*). Another driving force behind such studies is the need for infrastructure maintenance, tracking urban development and various traffic safety studies. These challenges necessitate the examination of extensive geographical regions, making it crucial to develop automated change detection techniques to reduce the manual effort and time required for data collection and processing.

Change detection in aerial and satellite images have numerous applications. In (*8*), the applications of change detection were discussed in general terms including areas such as civil infrastructure, driver assistance systems, and medical diagnosis and treatment which are outside of remote sensing. The objective of change detection in remote sensing is to (a) determine the nature of the change when feasible, (b) ascertain the spatial position of change observed by comparing two or more sets of imagery from different dates, and (c) measure the extent of the change (*11*). Therefore, crosswalk changes occurring due to modified infrastructure, modified position or number of crosswalks can be studied in a spatiotemporal aspect using remote sensing data and GIS.

Therefore, the primary objective of this research is to explore the creation of automated tools that employ computer vision and machine learning-based object detection models to identify changes in roadway elements, particularly crosswalks, in this paper. This objective can be reformulated as identifying the vector points that indicate the new or modified crosswalks in the new image (time instance $t_2$) compared to the old image (time instance $t_1$). The study will serve as a case study centering on the development of an automated GIS-based change detection model specifically tailored to detect and extract crosswalk changes from high-resolution aerial images. The goals of this study are to identify and extract a) crosswalks that were missing from the previous detections due to image occlusions, poor image resolution or faded markings, b) new crosswalks that have been built on new and existing roadways, and c) old crosswalks that have been improved or modified on existing roadways. In this study, the manual implementation of the framework on Orange County is shown. Also, a focus is given to the development of an automated change detection model where the existing crosswalk detector is utilized to extract all the crosswalk changes in Seminole and Osceola Counties.

This type of research holds significant importance for transportation agencies, serving multiple essential purposes in the context of roadway infrastructure maintenance, including a) the identification and tracking of new crosswalks that have been built overtime which are not part of the existing inventory, b) the identification of changes in crosswalks on both state and local roads, and c) the identification of previously missed crosswalks due to factors like low image resolution and occlusion. Another important contribution of this study is that it enables swift and frequent updates to the crosswalk inventory, which is an extremely critical pedestrian facility, eliminating errors introduced during manual processing methods. Keeping the crosswalk inventory up to date is crucial for various traffic operations and safety studies. By integrating the automatically extracted crosswalk change data with crash and traffic data, policymakers and roadway users can gain valuable insights and guidance in mapping crash patterns and crash trends. This will provide valuable information to understand the influence spatial relocation of crosswalk features have on crash occurrence.

## LITERATURE REVIEW
Several algorithms have been devised for remote sensing change detection. Over time, researchers like (*4*) and (*12*) have introduced a considerable variety of change detection techniques for remote sensing images and have compiled or categorized them from various perspectives. (*4*) grouped change detection methods into direct comparison and classification comparison whereas (*13*) expanded these categories by proposing object-based, pixel-based, and feature based change detection methods. (*12*) conducted a classification of change detection methods into seven types: advanced models, arithmetic operations, integration with Geographic Information Systems (GIS), comparison based on classification, transformation techniques, visual analysis, and miscellaneous methods.

Change detection approaches can be broadly grouped into two categories: bi-temporal change detection and temporal trajectory analysis. Bi-temporal change detection, which is commonly used in change detection, measures changes by comparing two dates whereas temporal trajectory measures changes in a continuous period by observing the progress over that period using profiles or curves of multitemporal data (*9*). Three methods are associated with bi-temporal change detection. These methods encompass comparing distinct data sources directly (direct comparison), comparing extracted information (post analysis comparison), and detecting changes by integrating all data sources into a standardized model (uniform modelling). On the other hand, temporal trajectory analysis involves two methods: disintegrated bi-temporal change detection, and extended time-series (*9*). Some notable change detection methods outlined by (*9*) are visual analysis, model method, image classification, object-oriented method, direct comparison, hybrid methods, and time-series analysis.

Visual analysis is used to detect changes by visually inspecting and identifying changes in bi-temporal images. This manual process is prone to errors and time consuming. The model method, on the other hand, is tailored to specific applications and is designed to understand the fundamental principles of application-related problems, using changes as the central elements to formulate mathematical models. The fundamental approach involves employing standardized models that incorporate all procedures and methods. One main challenge for this method is the availability of model parameters due to complexity of application problems (*9*). Hybrid methods apply two or more of the mentioned change detection methods. Although this method produces better change detection results, selection of hybrid methods for specific cases is difficult and confusing. The time series analysis method uses images with high temporal resolution but low spatial resolution for temporal trajectory analysis. Due to the low spatial resolution, analysis is usually restricted to vegetation and land cover changes. Vegetation indices like Enhanced Vegetation Index (EVI), Normalized Difference Vegetation Index (NDVI), Soil-Adjusted Vegetation Index (SAVI), and Advanced Vegetation Index (AVI) of given areas are used for estimating changes. The object-oriented change detection method involves identifying changes between corresponding objects extracted from multi-temporal images using image segmentation and other feature extraction algorithms. As this method directly compares the extracted objects, it is non-reactive to data noises and distortions in geometry. However, it may be sensitive to data extraction techniques that are prone to errors. Nonetheless, it is essential to pay attention to the techniques employed for comparing objects and accurately identifying changes between them. Object-oriented method is commonly used to detect changes in man-made objects on high resolution images and utilized in updating urban data. In summary, change detection methods can be broadly categorized into two main types: Bi-temporal image change detection and temporal trajectory analysis. Temporal trajectory analysis further divides into two subcategories: Long time-series analysis and real-time image sequence analysis. Conversely, Bi-temporal methods encompass three distinct forms: Direct comparison, involving direct pixel or image feature comparison; Post-analysis comparison, performed after classification or object extraction; and uniform modeling, where detection methods or processes are compared systematically. Additionally, visual analysis entails identifying changes through visual inspection. Combining any of these methods results in a hybrid approach, which integrates different techniques for more comprehensive change detection.

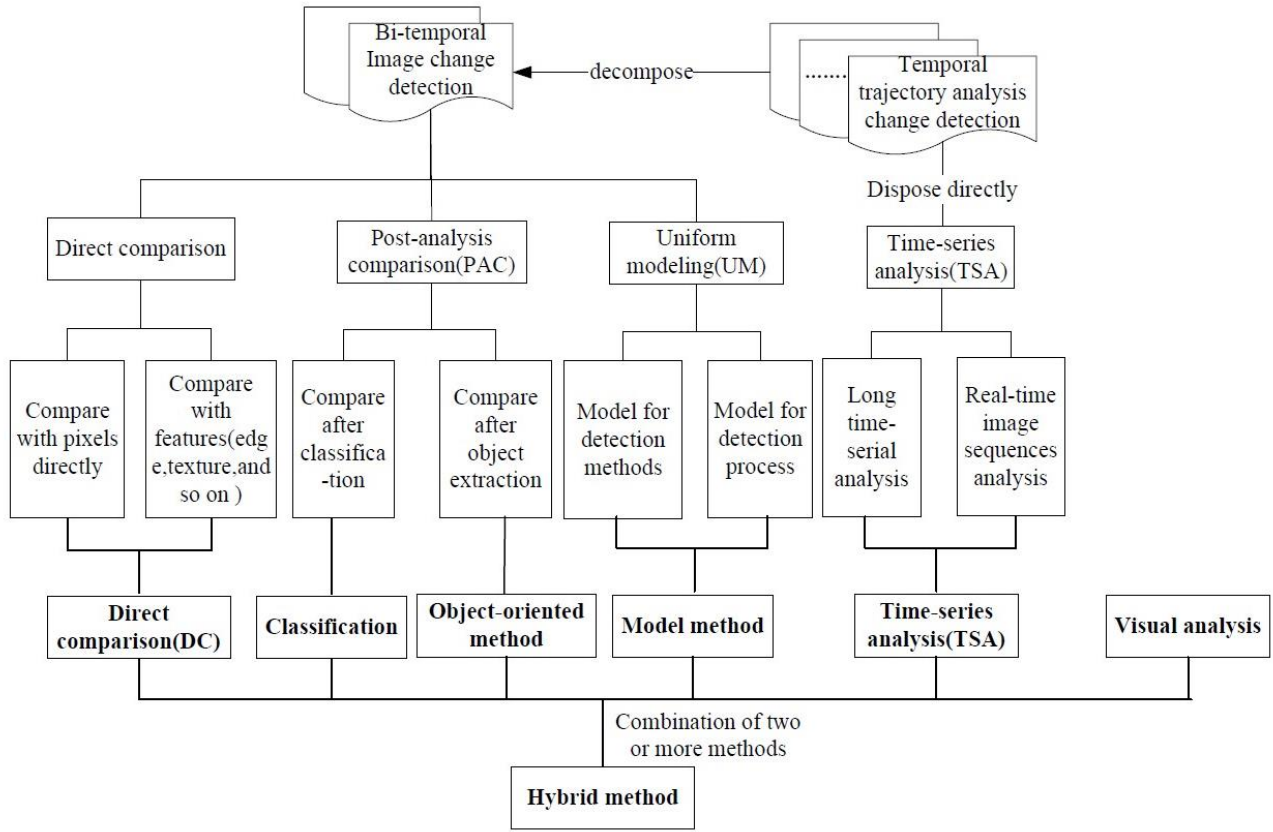

**Figure 1** Concept of change detection algorithms adapted from (*9*)

When conducting change detection, there are various crucial factors to consider. One factor to consider is the remote sensing system, such as spatial, spectral, radiometric, and temporal resolutions. Also, environmental factors, including atmospheric conditions, soil moisture, characteristics of natural and man-made phenological cycles, and tidal cycles should be considered (*14*). Additionally, one must also consider resources for image processing, the expertise and experience of the analyst, as well as time and cost limitations (*15*). It is noteworthy that most change detection methods developed by several scholars focused on pixel-based change detection methods. However, in this study, change detection was applied to features extracted from high resolution aerial images using an existing crosswalk detector model (*16*). GIS is playing a pivotal role in transportation studies with applications for planning, designing, decision-making, network analysis and monitoring of interruptions in transportation systems (*17,18*) in this current information driven society. The use of this technology in fuel and gas distribution utilities has in this present time expanded tremendously. GIS application components help to model geospatial information and processes that support fuel and gas distribution utility network and operations in the real world. Incorporating GIS into change detection has facilitated the seamless integration of source information, making the process more straightforward (*19*). Additionally, it has proven to be beneficial in effectively visualizing the detected changes. By integrating source data with quantitative information, GIS simplifies the extraction and assessment of change detection data. Geospatial application has proven to be the best tool for vector data studies and analysis. Researchers in transportation data science have employed diverse methodologies to extract roadway geometry data from aerial imagery using Artificial Intelligence (AI) techniques. For instance, in a recent study by (*20*), a YOLOv3 school zone detector was developed and implemented to identify school zones from aerial images. The present research extends this approach to create a statewide inventory of school zone pavement markings in Florida by automating the extraction process using a customized object detector algorithm. This innovative model significantly reduces the need for labor-intensive and error-prone manual detection methods. Its capabilities include identifying outdated and partially visible school zone markings, offering a valuable tool for various transportation analyses conducted by transportation agencies. However, it's worth noting that one limitation of the developed model lies in its inability to compare old and new school zones or update existing school zone inventories. Another study by (*21*), detected turning lane features such as left, right and center lanes from aerial images using Yolov5. Researchers in (*22*) extracted footpaths, sidewalks, and crosswalks from aerial imagery using semantic segmentation methods to create an inventory of these pedestrian facilities. Another researcher in (*23*) utilized deep learning approach to detect and extract crosswalks from satellite images. In this study, YOLO-based detector was developed to detect and extract these crosswalks. Moreover, (*16*) employed a YOLO-based object detector to extract crosswalks from aerial images. In that study, a developed crosswalk detector was utilized to identify various pavement markings in high-resolution aerial imagery spanning the entire state of Florida. Through a sensitivity analysis, crosswalks were categorized into midblock crosswalks, signalized intersection crosswalks, and driveway crosswalks. The outcomes of the research contributed to the creation of a comprehensive crosswalk inventory, a valuable resource for research in roadway safety and transportation policy. Despite the success in automating crosswalk inventory creation, a notable limitation is its inability to provide regular updates when new crosswalks are established over time. Building upon this, the ongoing research aims to enhance the existing crosswalk detector algorithm by developing an automated crosswalk change detection system. The major contributions of this study are highlighted as follows: 1. Automated identification and extraction of georeferenced new crosswalks, 2. Identification and extraction of modified old crosswalks, and 3. Identification and extraction of crosswalks missed during initial detections due to occlusions and poor image resolutions. This advancement is envisioned to facilitate regular and efficient updates to the existing inventory, eliminating redundant detections.

Elements that are detected when using direct comparison method are the basic image features, pixels or transformed image features. The primary components of the post-analysis comparison method for detection mainly consist of objects extracted from images. This study approaches change detection using a post-analysis comparison, specifically using the object-oriented method by creating a model that detect the changes from crosswalk features extracted from high resolution aerial images from two different time periods. This approach has the potential to eliminate potential data redundancy and losses during RCI data inventory update.

## STUDY AREA

The case study area chosen includes the counties of Orange, Seminole and Osceola, Florida. We will develop the framework using Orange County and use the developed model to identify the changes in Seminole and Osceola counties. Note that Orange County includes the City of Orlando and is depicted in **Figure 2a**. Covering an area of 1,003 square miles, the county shares borders with Brevard County to the east, Seminole and Lake Counties to the north, Lake County to the west, and Osceola County to the south (*24*) with more than 2,600 miles of roadway system (*25*). In the county, the land-surface elevations do not exceed 250 feet, and the geography exhibits notable variations from the western to the eastern parts. The study area mainly comprises light commercial, residential, and open irrigated spaces, including golf courses and school grounds (*26*). As of 2020, the estimated population of Orange County exceeds 1.42 million, as reported by the U.S. Census Bureau (*24*). Seminole County, Florida covers an area of 309.5 square miles, ranking as the 64th largest county in the state based on total area (*24*). The county's roadway system encompasses approximately 1900 miles of roads, comprising state, county, and city roads (**Figure 2b**) (*27*). Seminole County shares its borders with Orange County, Volusia County, Brevard County, and Lake County, all located in Florida. As per the 2020 population census (*24*), the county's population stands at 470,856. Osceola County, Florida spans an area of 1,327.7 square miles, making it the sixth largest county in the state in terms of total area (*24*) with over 600 miles of roadway length (*28*). The county shares its borders with Okeechobee County, Lake County, Polk County, Brevard County, Indian River County, Highlands County, and Orange County, all situated within Florida and is home to a population of 388,656 residents (**Figure 2c**) (*24*).

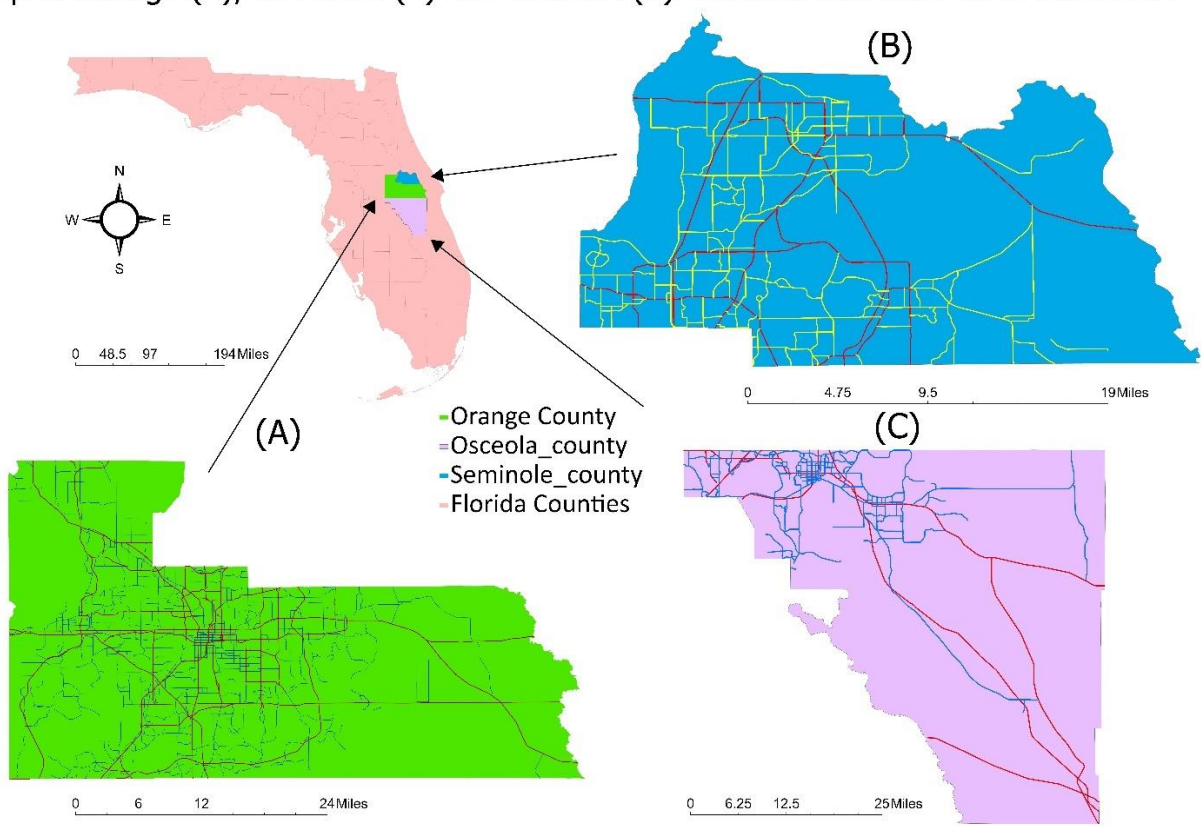

**Figure 2** Map of Studied Counties in Florida with the roadway network: (a) Orange county, (b) Seminole county, and (c) Osceola county

## MATERIALS AND METHODOLOGY
The selection of the data collection method for roadway inventory depends on various factors, including data collection time (e.g., collection, reduction, and processing), cost (i.e., collection and reduction expenses), accuracy, safety considerations, and data storage requirements. In this study, our objective was to develop a change detection framework and model specifically designed to identify changes in crosswalk features extracted from high-resolution aerial images in selected three counties of Florida. This study was conducted in two parts: Firstly, a manual change detection method was employed to obtain the crosswalk changes in Orange County, Florida between 2019 and 2021. Afterwards, an automated change detection method was developed to extract crosswalk changes in Seminole and Osceola counties, Florida, between 2018 and 2021, and between 2019 and 2020, respectively. In the manual change detection process, the crosswalk changes were identified by superimposing the detected crosswalks on the old and new image data to manually identify and remove newer crosswalks that are in closest proximity (36 ft.) to older crosswalks using visual inspections. However, in the automated method, a GIS-based change detection model was developed to automatically extract crosswalk changes using variables such as old and new images, csv files and their coordinate systems, output shapefile names, and roadway data. The developed automated model in this study can locate the areas where a crosswalk change has occurred.

*Data Description*
The Florida Department of Transportation's (FDOT) Surveying and Mapping Office maintains an archive of aerial imagery. These images, covering all 67 counties in Florida, are well-indexed and georeferenced, spanning several years (*29*). The image filenames contain the three-letter county code, the year the image was taken, and the tile number. For this study, high-resolution aerial images were sourced from the Florida Aerial Photo Look-Up System (APLUS), an accessible archive managed by the FDOT Surveying and Mapping Office, available to the general public (*30*). While a limited number of photos can be downloaded from the website, acquiring large datasets that cover entire counties or states requires sending a request and letter or providing an external drive with sufficient storage capacity. For this research, the most recent high-resolution images, dated December 2022, were collected for Orange (2021), Seminole (2021), and Osceola (2020) counties, totaling 120 GB in size. Additionally, older images for the mentioned counties, including Orange (2019), Osceola (2019), and Seminole (2018), were also utilized. **Figure 3** visualizes some changes observed between old and new high resolution aerial images utilized in this study.

The aerial images used in this study show varying resolutions, ranging from 1.5 feet to 0.25 feet. The majority of images have a resolution of 0.5 feet per pixel (0.15 meters per pixel) and are presented in a size range of 10,000 × 10,000 to 20,000 × 20,000 pixels, featuring a 3-band (RGB) image format (**Figure 3**). However, the exact resolution may differ depending on the specific county. Furthermore, these images are provided in MrSID format, enabling GIS projection on a map (*29*). Additionally, another source of GIS data is available through the FDOT Transportation Data and Analytics Office (*28*). This information is categorized into four main groups: (a) data on designated roadways, (b) data on roadway features, (c) data on traffic, and (d) data on bicyclists and pedestrians.

This study primarily focuses on crosswalks located on roads under county or city jurisdiction, as well as those present on roads within the state highway system. Interstate roadways were excluded from the state road data, and all centerlines from the county and city-controlled road shapefile were combined. In this study, state highway system roads are referred to as ON System Roadways or state roads, while county- or city-controlled roads are described as OFF System Roadways or local roads, following the classification by FDOT. It should be noted that while FDOT's GIS data can provide several geometric data points necessary for mobility and safety evaluations, it lacks up-to-date information on the positions of crosswalks on both state and local roadways. Consequently, the main objective of this project is to employ a change detection framework to extract, compile, and update the existing inventory of crosswalk markers on both state and local roadways in Orange County, Florida. This approach was automated and subsequently utilized to extract crosswalk changes in Seminole and Osceola Counties. In summary, this methodology involved several steps to analyze changes in high-resolution aerial images. Initially, both old and new versions of the aerial images were obtained and preprocessed using the masking model. This masking model which effectively isolated the regions of interest, mainly focused on extracting the roadways from the images. Following the preprocessing step, metadata from the masked images, which included important information such as spatial information and image properties were extracted. This metadata together with the images were then fed into the detection script specifically designed to identify crosswalk points within the roadways. The detection script utilized advanced computer vision and machine learning techniques to accurately locate and extract crosswalks from the images. The extracted crosswalk points were passed through a change detection algorithm, which compared the crosswalks between the old and new images to identify any significant changes. This change detection process was crucial in identifying variations such as new crosswalks, modified crosswalks, or removed crosswalk features. Finally, the output from the change detection algorithm provided a set of crosswalk data indicating the detected changes. To ensure accuracy and reliability, a post-processing script that systematically screened the data to remove any duplicate detections and false positives was developed. This post-processing step was essential in refining the final crosswalk data, resulting in a more reliable dataset for further analysis and interpretation.

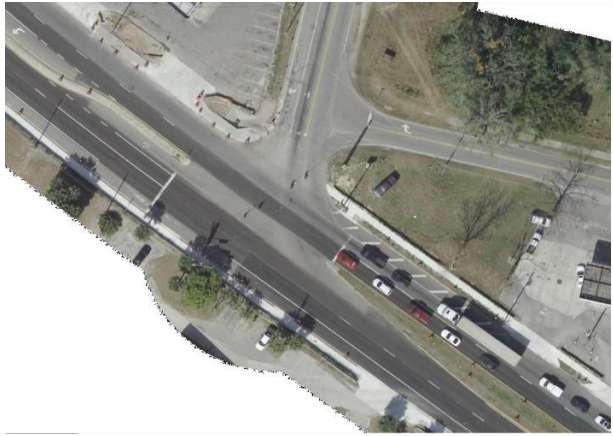 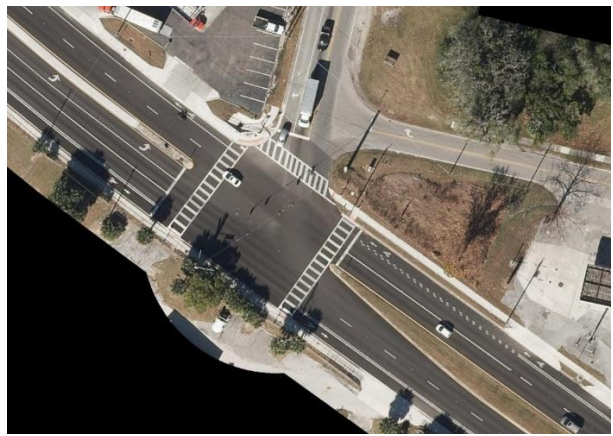

**(a)** **(b)**

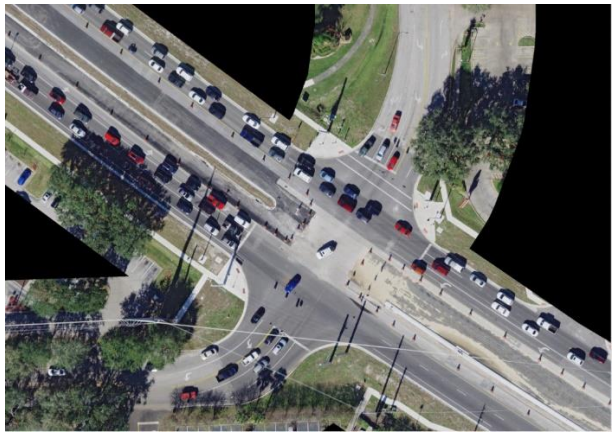 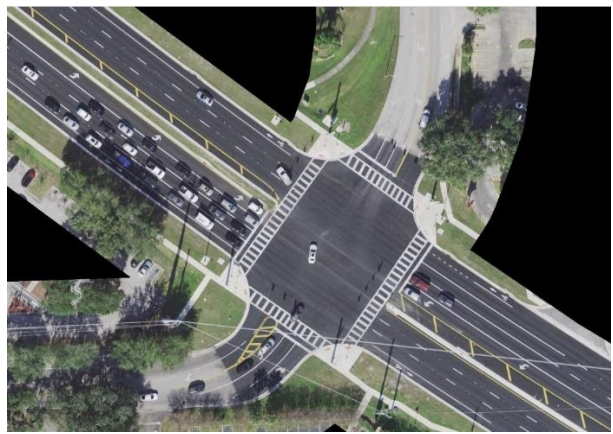

**(c)** **(d)**

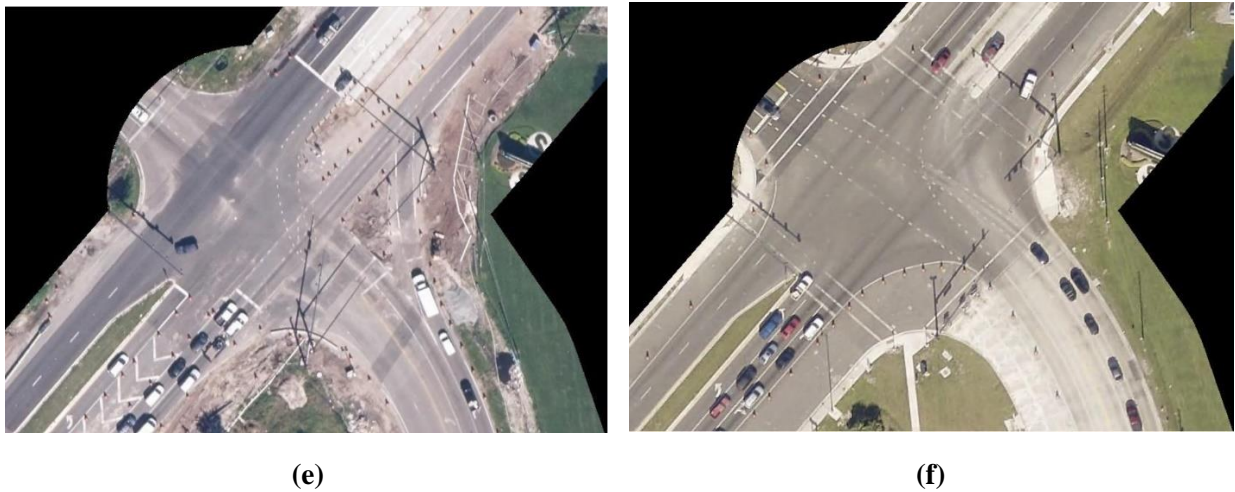

**(e)** **(f)**

**Figure 3** Old and new aerial images of study area: (a) 2019 image Orange county, (b) 2021 image Orange county, (c) 2019 image Osceola county, (d) 2020 image Osceola county, (e) 2018 image Seminole county, and (f) 2021 image Seminole county.

*Pre-processing*
Preprocessing is a required step because of the volume of data and the complexity of the object recognition process (*31*). Our technique generally selects and discards any images that do not cross a roadway centerline, and then it masks out any pixels that are not surrounded by a buffer zone. This method reduced the amount of images from 90,000 to 30,000, and the image masking model disallowed items that were closer than 100 feet to state and municipal roads (*20*). The highway shapefile that was utilized was buffered to create polygons with dissolved overlapping borders before the images were masked. Aerial images were iterated using this layer as a guide, and any crossing areas were clipped. Pixels that fell outside the limit of the reference layer were eliminated during the masking step. A single raster file was created by mosaicking the clipped images, which had less pixels as a result. For any type of raster analysis or data processing, this new file was smaller and simpler to handle in a mosaic format.

The preprocessing approach is thoroughly depicted in **Figure 4a**. Using the ArcGIS Pro software, all of the images from the selected counties (n = around 90,000) were first imported into a mosaic dataset. Mosaic databases were used to organize and display a number of georeferenced images (*32*). Additionally, mosaic datasets make it possible to choose picture tiles based on location by intersecting them with extra geocoded vector data. For instance, by choosing and extracting individual photos that make up a section of the roadway centerline, a subset collection of the photographs (n=30,000) was produced. Furthermore, the ArcGIS Pro modelbuilder interface was used to construct an automated photo masking tool.

**Figure 4b** illustrates how the program continually scans through a folder of photos, adds a mask based on a 100-foot buffer around the road's centerlines to each image, and then saves the masked images as JPG files for the object identification technique. This process was used to mask all the aerial images during pre-processing.

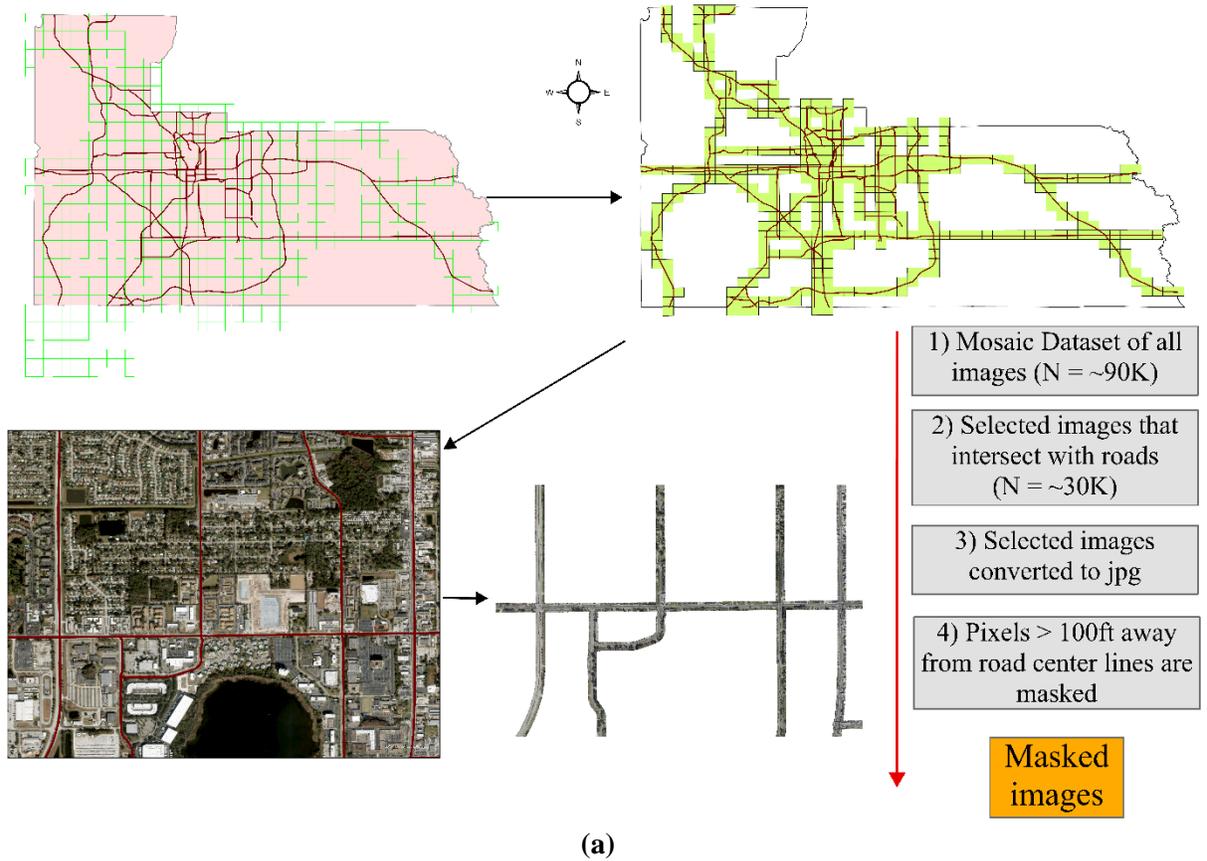

**(a)**

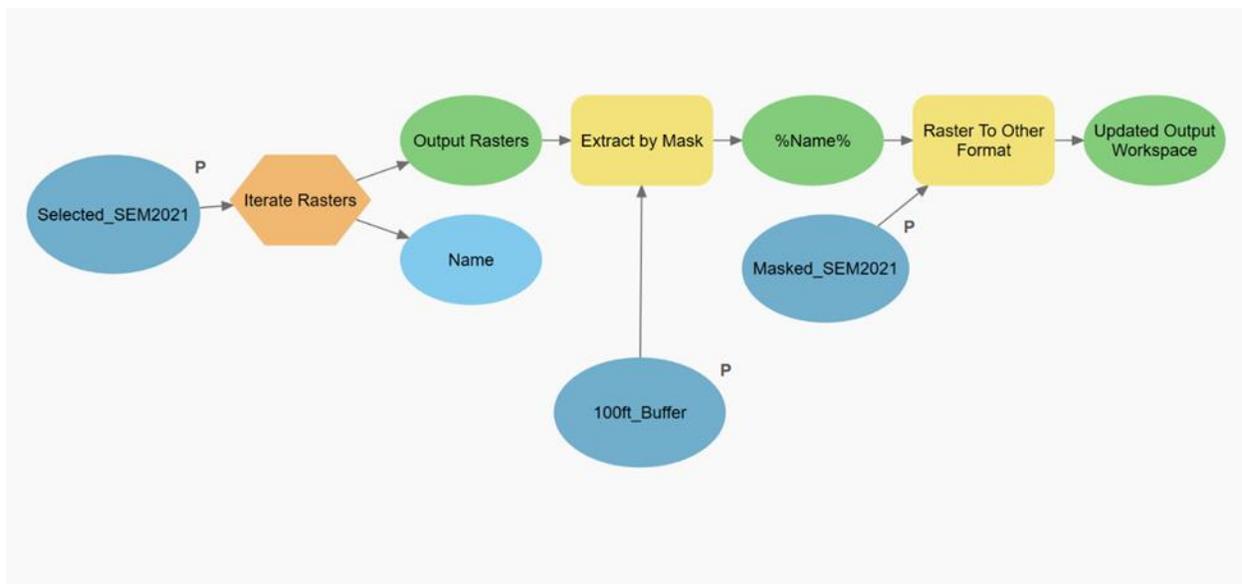

**(b)**

Antwi, Takyi, Karaer, Ozguven, Kimollo, Moses, Dulebenets, and Sando

**Figure 4** (a) Preprocessing approach, and (b) automated image masking model used in the fourth step of preprocessing adopted from (*20*).

The positions of each image were then specified precisely in a metadata database. Remember that object detection algorithms can only discern which pixels in a picture are covering an object. The position of every pixel in masked photographs must be precisely known to map the detected crosswalks. The exact corner placements of each image may be determined in feet, even if the clarity and size of the photographs vary throughout different counties. This table is read by the object identification model, which then makes use of the size information, scales the picture as necessary, and instantly pinpoints the locations of any found crosswalks.

*Existing YOLOv2 Crosswalk Detection Model*
The YOLOv2 (You Only Look Once) theory is the foundation of the current crosswalk detection model developed by (*16*) and implemented in 2022. Please note that every version of YOLO is designed for detecting multiple objects in typical surface-level images. This design has proven highly effective for real-time applications like automated driving. However, among the available YOLO versions during the development of the model, YOLOv2 was specifically chosen in this study because it was used to detect single class objects from large input image sizes after customization. Compared to the other versions at the time of the study (*16*), YOLOv2 performed better in this regard since the goal was to efficiently analyze high-resolution aerial images across the entire State of Florida to ensure accurate detection of crosswalks with diverse pavement markings and varying width and height ratios. The existing object detector model was trained through a transfer learning approach in MATLAB v.2020a using manually-labelled 4,658 crosswalk features mostly collected from Alachua, Duval, and Miami-Dade Counties' high-resolution images. While most of the training data comprised of crosswalks from Alachua (1360), Duval (756), and Miami-Dade (669) counties, a comprehensive representation included crosswalks from 28 counties. Out of this, 70% was used to train the model, 15% was used to test the model, while the remaining 15% was used to validate the model during training process. During the training data creation process, bounding boxes were drawn around crosswalk features identified on the high-resolution images. Therefore, the training data was made up of the features that were extracted from the images into image chips that have metadata containing the bounding box coordinates and feature descriptions. The model was trained at 30 epochs, 128 batch size, 0.0004 L2 Regularization value. The precision-recall graph of the existing crosswalk model is shown in **Figure 5a**. Crosswalks are identified by the model as points with a minimum confidence level of 0.4. The model's performance is evaluated using the three criteria of completeness, accuracy, and quality. Following their first usage for highway extraction in (*33*) and (*34*), these measures are often employed for performance evaluation of related models (*35,36*). Two different datasets were collected and utilized for evaluating the model's performance. The ground truth (*GT*) crosswalk datasets used to evaluate the performance of the model were made up of manually labeled 1,272 crosswalk points and 2,312 Open Street Map (OSM) crosswalk points from Leon County, FL. This is recorded in the **Table 1**. The manually labelled *GT* data was obtained by visually inspecting each visible crosswalk while using a masked image as background and placing a point on the observed crosswalks from the image. From the evaluations, the model performed better when output crosswalk detections were compared to the manually labelled GT data than when compared with the obtained OSM data.

The following are determination rules required to calculate the performance evaluation metrics:

i. <u>Ground Truth</u> (*GT*): Number of GT crosswalk points.

    ii.    <u>Model</u> (*M*): Number of Model detected crosswalk points.
    iii.    <u>False Negative</u> (*FN*): # of GT crosswalk points with no M crosswalks within 30ft. perimeter.
    iv.    <u>False Positive</u> (*FP*): # of M crosswalk with no GT crosswalk point within 30 ft. perimeter.
    v.    <u>True Positive</u> (*TP*): # of M crosswalk with GT crosswalk point within 30ft. perimeter.

Performance evaluation metrics:

Completeness $= \frac{GT-FN}{GT} * 100\%$, true detection rate among GT crosswalk (recall)

Correctness $= \frac{M-FP}{M} * 100\%$, True detection rate among M crosswalk (precision)

Quality $= \frac{GT-FN}{GT+FP} * 100\%$, True detection among M crosswalk plus the undetected GT crosswalk (Intersection over Union: IoU)

**Table 1**: Model performance evaluation compared to OSM adapted from (*16*)

| Ground truth (GT) (*n* = 1272) | Model (M) (*n* = 1316) | OSM (*n* = 2312) |
|---|---|---|
| TP | 1092 | 989 |
| FN | 180 | 283 |
| FP | 149 | 1208 |
| Recall | **85.9%** | **77.8%** |
| Precision | **88.7%** | **52.2%** |
| Quality | **76.9%** | **39.8%** |

    The generated model's accuracy, as determined by performance assessment measures, is 85.9%. The model accurately recognizes crosswalks with an accuracy score of 88.7% and a quality value of 76.9%. A CSV output containing the coordinate coordinates of the discovered crosswalks is produced after the masked pictures have been run through the detection model. The earlier and more recent photos of the Orange, Seminole, and Osceola Counties were obtained using the same procedure, which was repeated.

*Change Detection for Orange County*
The crosswalk detection algorithm was run on both the old and new high resolution aerial images. For additional investigation, detections were extracted into a CSV file. In order to start the change detection analysis, the images together with the 2019 and 2021 detection CSV files of the crosswalks in Orange County were imported into ArcGIS Pro. To facilitate data processing, the CSV was included as a shapefile. For the 2019 and 2021 crosswalk shapefiles, different symbologies were used, and these were superimposed on the image data for easy identification. The crosswalks in the 2021 image that are closest to those in the 2019 image—at a search distance of 36 ft—were extracted for analysis. The features that were extracted were used to generate a new layer. The Orange County state roadway was buffered for a distance of 100 feet, and that distance was used to obtain the crosswalks that are located there. The procedure used for the manual crosswalk change detection technique is shown in **Figure 5b**.

*Antwi, Takyi, Karaer, Ozguven, Kimollo, Moses, Dulebenets, and Sando*

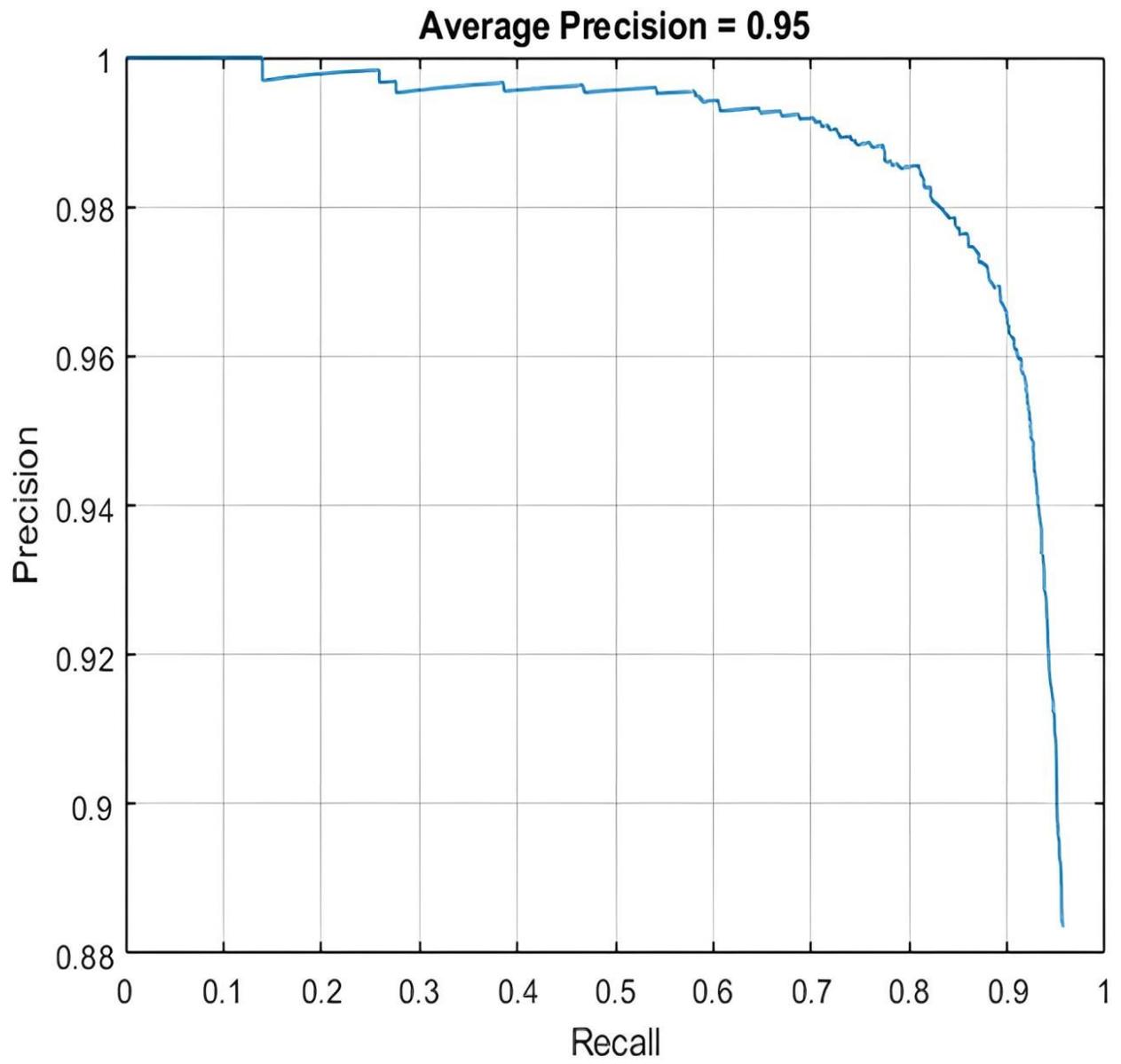

(a)

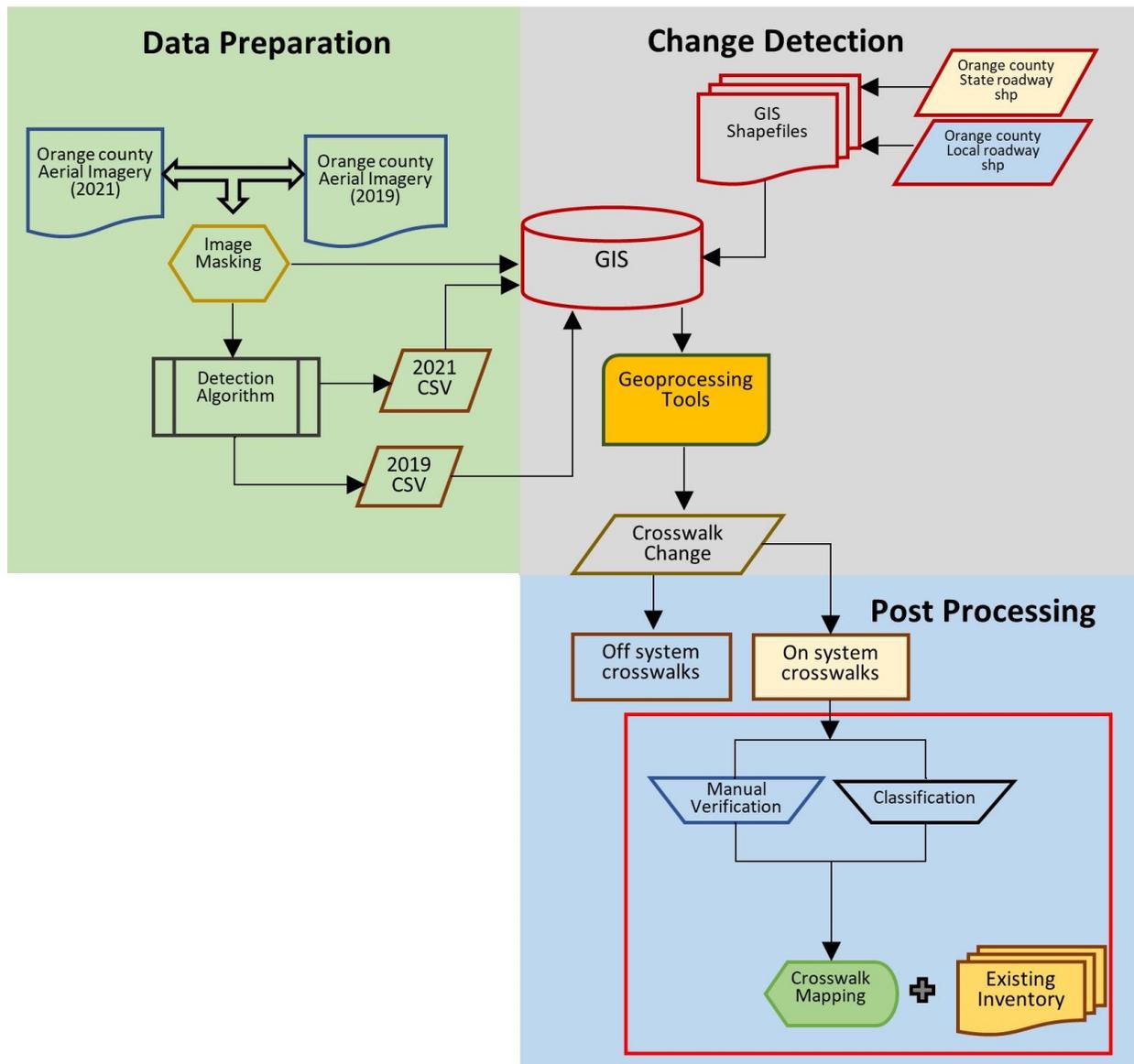

(b)

**Figure 5**: (a) Precision-recall graph of crosswalk detector model and (b) description of the workflow undertaken for the crosswalk change detection process and the manual verification of changes (post-processing).

*Automatic Crosswalk Change Detector*
The developed crosswalk detection algorithm was run on both the old and new high resolution aerial images for Seminole and Osceola Counties. For additional investigation, detections were extracted into a CSV file. Two variables were created for each of the CSV files collected from the old and new image detections in order to start building the model. The source folder holding the CSV file was set as one variable, while the image's coordinate system was specified as the other. The model imports the CSV files and converts them into shapefiles using the specified coordinate system. This step is concurrently performed for the older image CSV. Then, the new crosswalks that are more than 36 feet apart from the previous crosswalks are selected using the imported selection tool within the model. This was accomplished by modifying the selection parameters to perform an inverse selection of new image crosswalks that are within a distance of 36 feet of the old image crosswalks. This assumes that any newly constructed or modified crosswalks will be at least 36 feet away from already existing crosswalks. Therefore, except for crosswalks that were missed in the previous picture owing to conditions like occlusion, poor resolution, or faded markings, all crosswalks on the new image that are within a distance of 36 feet of the identified crosswalks on the old image will be disregarded. As a result, wherever these crosswalks are found on the new picture, they will also be included. The newly generated layer and the number of crosswalk changes are included in the output from the selection. A folder directory was added to the model to export the output features. The crosswalk modifications for both state and local roadways are included in this exported file. The state road shapefile was now used as a masking feature in another selection to extract crosswalks on state highways. The updated result was placed in a folder. The CSV file folders, coordinate systems, highway shapefile, and output file names and directories were the parameters used to generate the model. The automated system was used to extract the Seminole and Osceola crosswalk modifications (**Figure 6a**).

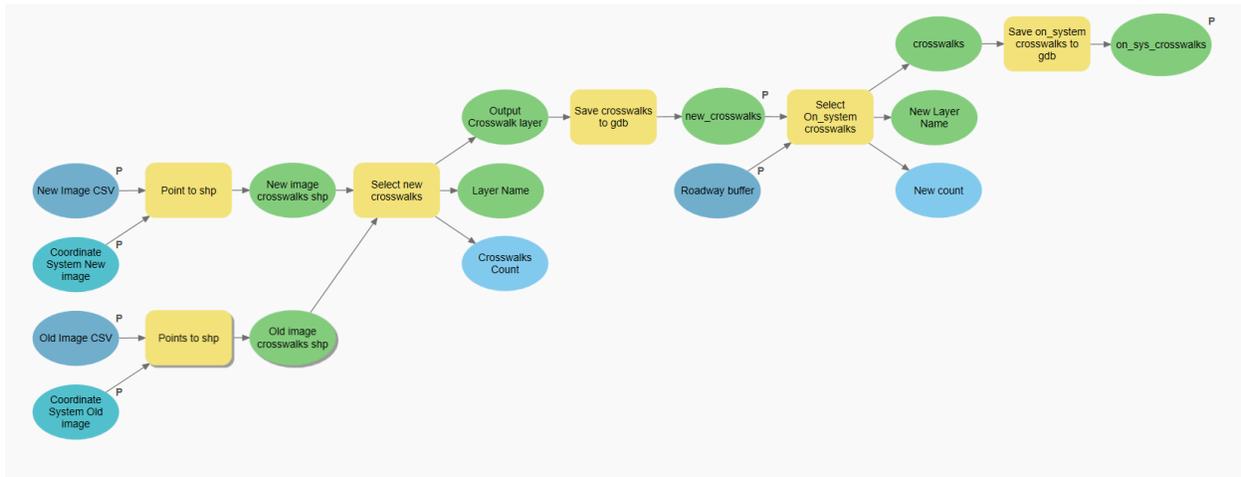

(a)

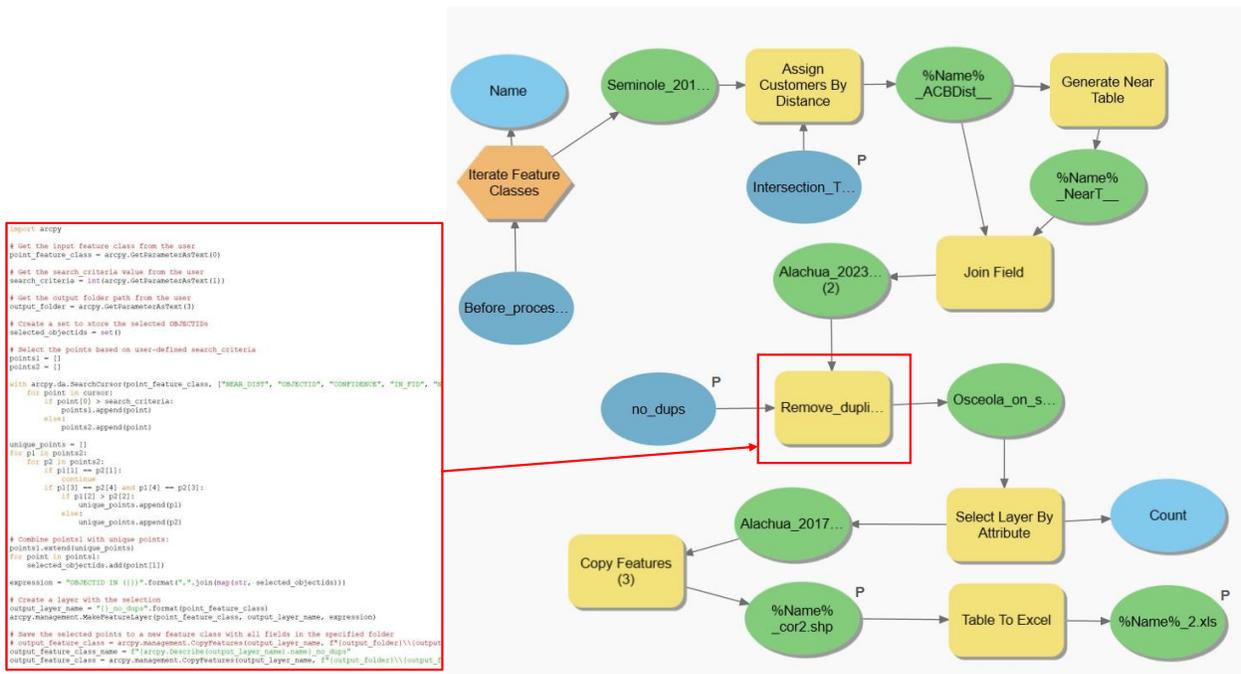

(b)

**Figure 6** (a) Framework of developed automatic crosswalk change detection model and (b) post-processing script to handle false positives and duplicate detections.

*Post-processing*

The extracted crosswalks in selected counties were checked by physically inspecting each of the chosen crosswalks to eliminate false positives and categorize crosswalks according to the roadway condition (new roadway or modified existing roadway). The discovered changes were manually examined and divided into new and pre-existing crosswalks as part of the post-processing procedure. Since the initial detections contained false positives and duplicate detections, a duplicate detection removal algorithm was created together with a two-condition false positive removal script to handle these errors. The duplicate detection method like non-maximum suppression was applied to remove the duplicates. Here, the algorithm considers any 2 points within 24ft (user – specified) and closest to each other as a duplicate and removes the crosswalk point with lowest confidence level. Duplicate removal algorithm can always be altered to increase distance threshold. False positives were also handled using a script to select and discard all crosswalks that meet 2 conditions; having lower confidence level <50 and more than 90ft. away from intersections (**Figure 6b**). The new crosswalks include those that have been constructed on either brand-new or pre-existing roads. Old crosswalks that were missed in prior detection processes because of issues like inadequate picture quality, faded markings, or occlusions, on the other hand, are considered modified or improved crosswalks. These are crosswalks that were previously on existing highways but have since been upgraded or changed. For example, the changes between the crosswalks in Orange County from 2019 to 2021 are depicted in **Table 2**. The yellow points represent newly identified crosswalks from a 2021 image whereas the red points represent crosswalks that were recognized in a 2019 image.

**Table 2** Examples of differences between 2019 and 2021 images in Orange County (red points: old crosswalks, yellow points: new crosswalks)

| Classification | 2019 | 2021 |
|---|---|---|
| New crosswalk | 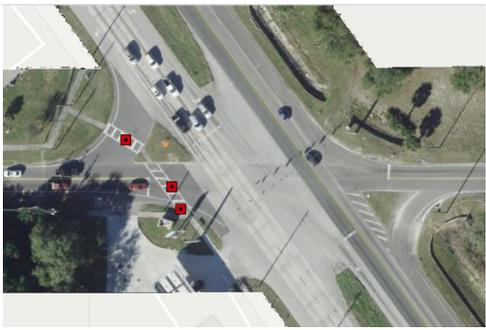 | 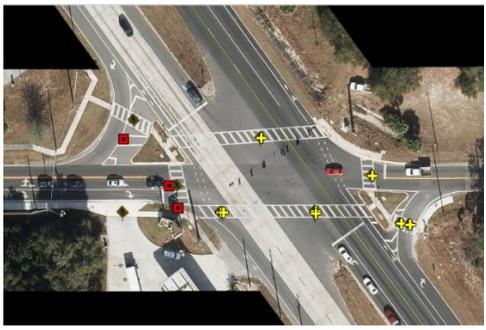 |
| New crosswalk | 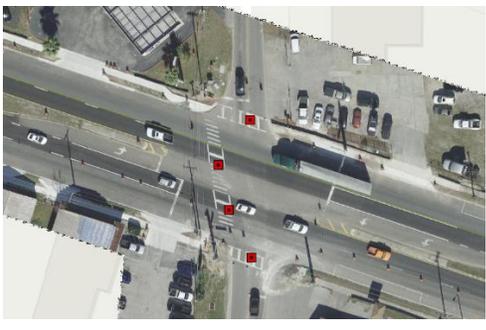 | 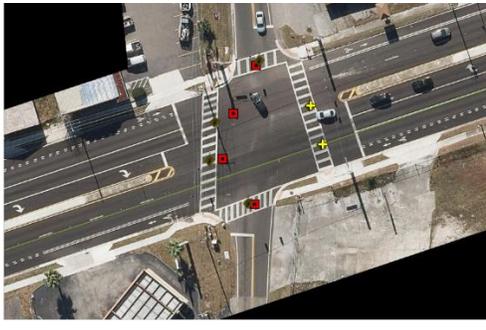 |

*Antwi, Takyi, Karaer, Ozguven, Kimollo, Moses, Dulebenets, and Sando*

| | | |
|---|---|---|
| New crosswalk | 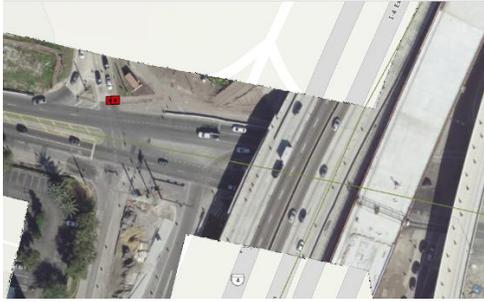 | 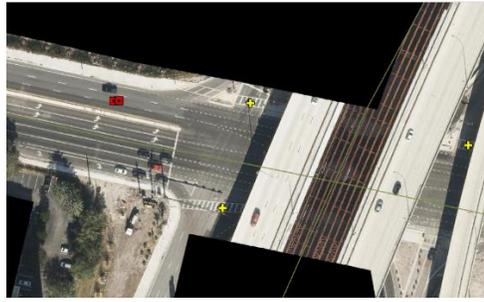 |
| New and modified crosswalks | 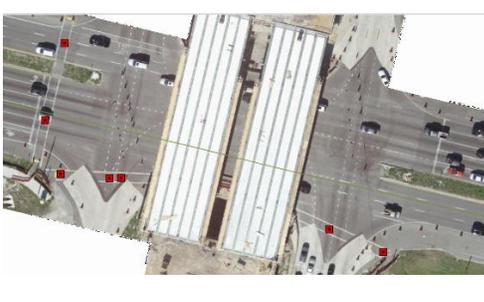 | 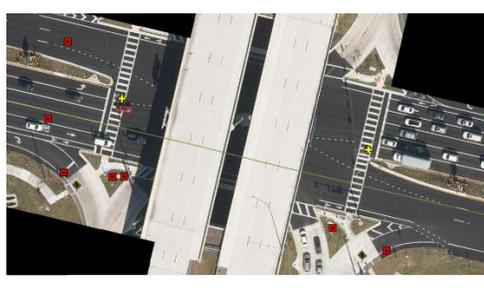 |
| New crosswalk | 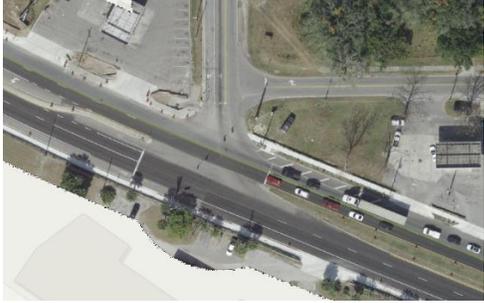 | 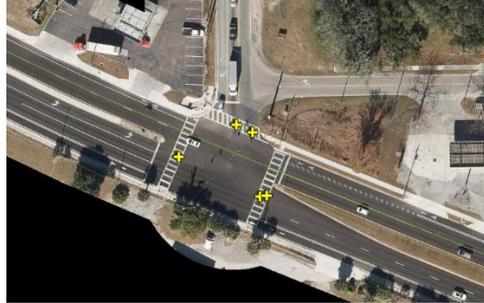 |
| New crosswalk | 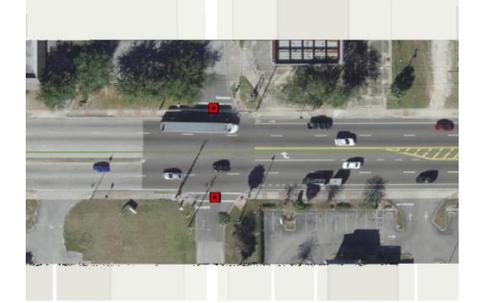 | 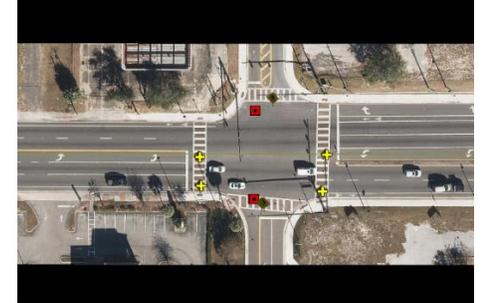 |

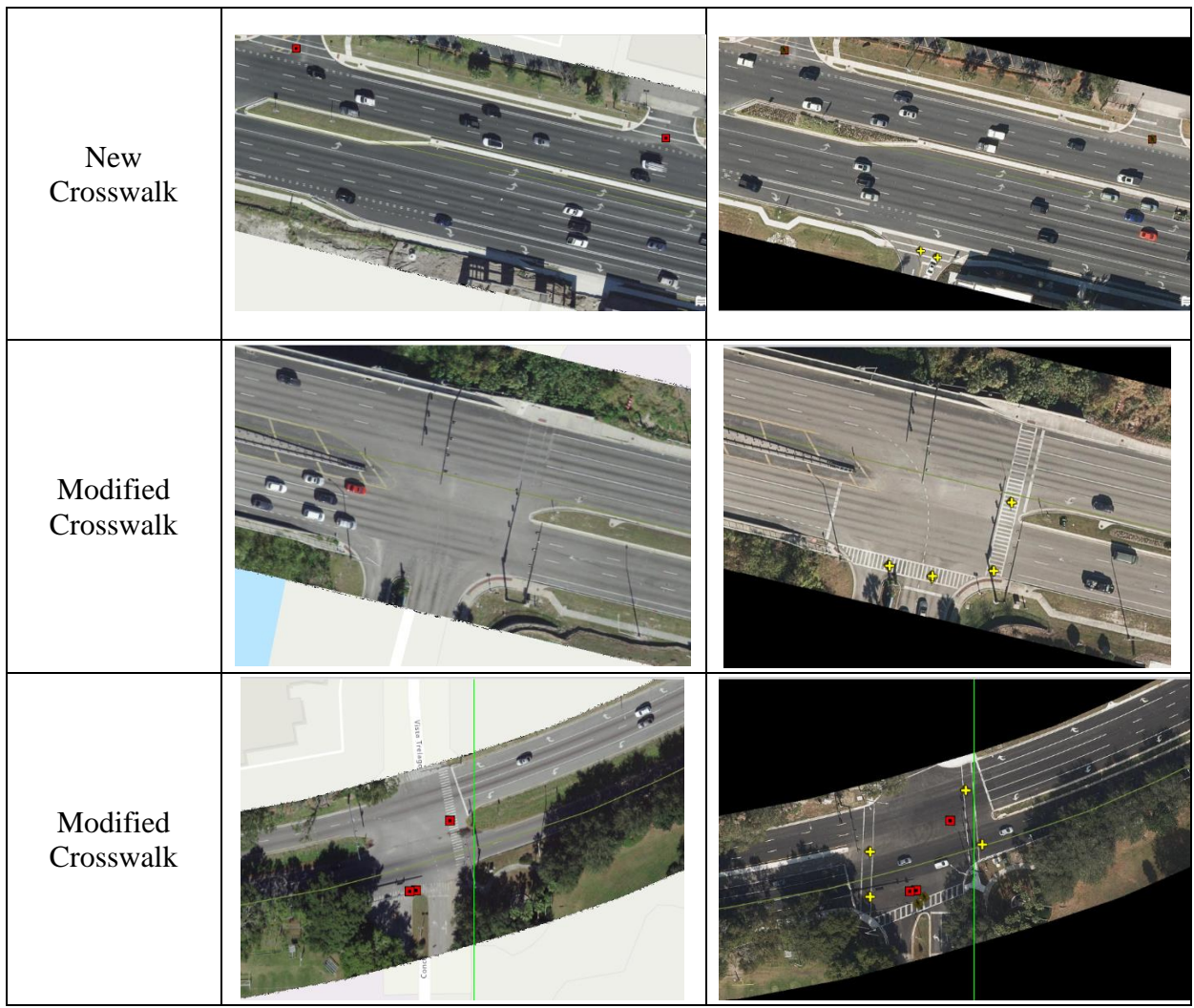

## RESULTS AND DISCUSSIONS

### *Case Study: Detected crosswalk changes in Orange County*
The developed framework for detecting changes in crosswalks offers a method for comparing crosswalks' current and previous states, tracking changes in crosswalks, and taking appropriate action in response to those changes. The new crosswalks include those that have been built on new roadways or others that have been built on existing roadways. On the other hand, modified or improved crosswalks include a) old crosswalks on existing roadways that have been modified or improved, and b) old crosswalks that were missed in previous detection process due to factors like as poor image resolution, faded markings, or occlusions. The differences or modifications found can be utilized to update the current inventory and support the scheduling of maintenance tasks. In the first case study, the crosswalk modifications in Orange County were manually extracted and categorized following visual examination. As proof of concept, a complete dataset of Orange County's crosswalk changes was gathered and carefully reviewed. The masked images served as the background as the visible crosswalk markings were classified.

*Antwi, Takyi, Karaer, Ozguven, Kimollo, Moses, Dulebenets, and Sando*

The total number of detections observed before post-processing in the 2019 imagery was 12,847 while 2021 imagery was 12,190. Note that, the total number of detections include both true and false positives, as well as duplicate detections. The total number of observed crosswalk changes from 2019 to 2021 before post-processing was 2,094 which included crosswalks on both state and local roadways. Out of the 2,094 observed changes, 314 detections were observed on the state roadways. After manual verification and removing duplicates, the total number of observed crosswalk changes was 769 where 166 were found on the state roadways in Orange County. Out of these, 108 of them were new crosswalks that had been built whereas 58 of them were existing crosswalks that were modified or improved. This number also included crosswalks that were missed in the previous detection process due to factors such as poor image resolution, faded markings, or occlusions. The results from the crosswalk change detection were mapped for better visualization (**Figure 7**).

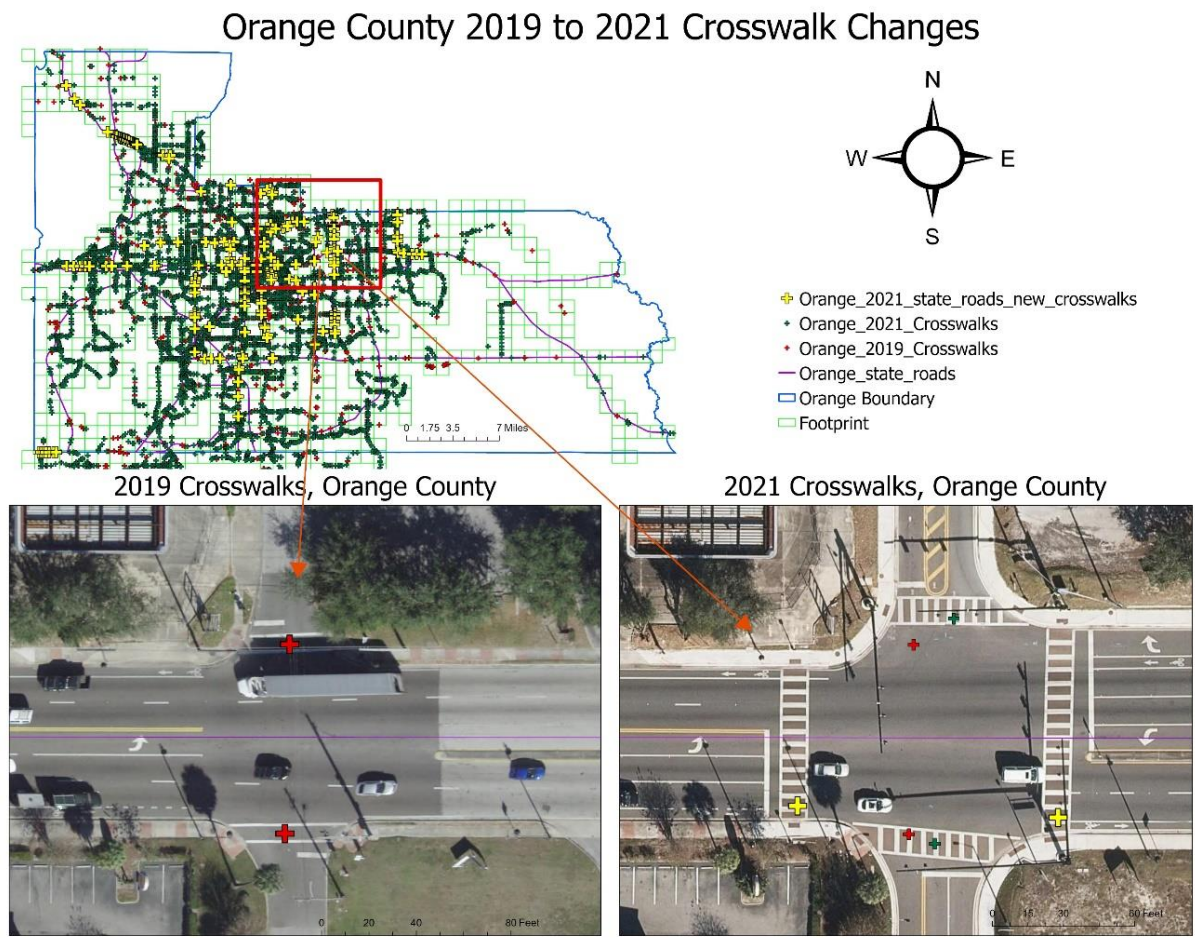

**Figure 7:** Map Showing Crosswalk Changes in Orange County between 2019 and 2021

*Detecting Crosswalk Changes in Seminole and Osceola Counties*
The developed change detection model was used to detect crosswalk changes in Seminole and Osceola Counties. The overall count of crosswalk changes identified in Seminole County between 2018 and 2021 was 1,040, with 340 of them located on the state roadway system. Similarly, in Osceola County, a total of 1,402 detected changes were observed from 2019 to 2020, with 344 of these changes found on the state roadway system. It is worth noting that this number also includes crosswalks that were not detected in the previous process due to factors such as poor image resolution, faded markings, or occlusions. After postprocessing, from the results, the new crosswalks observed on the state roads in Seminole County from 2018 to 2021 were 123, and Osceola County from 2019 to 2020 were 136. The results of the crosswalk change detection in Seminole and Osceola Counties were visualized and mapped for improved clarity (**Figure 8 and 9**). The identified differences or modifications can be used to update the existing inventory as well as to aid in planning maintenance activities.

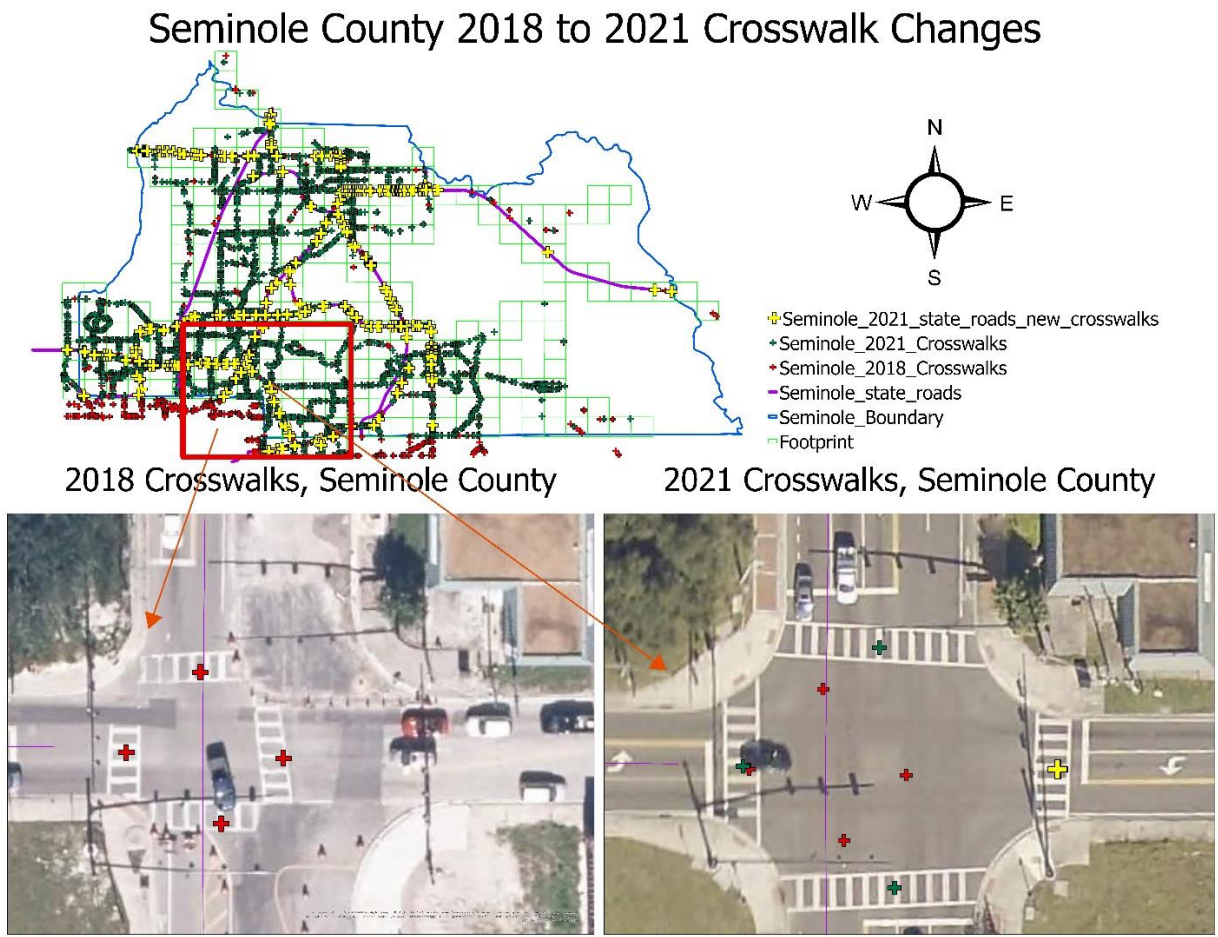

**Figure 8:** 2018 to 2021 Crosswalks Changes in Seminole County

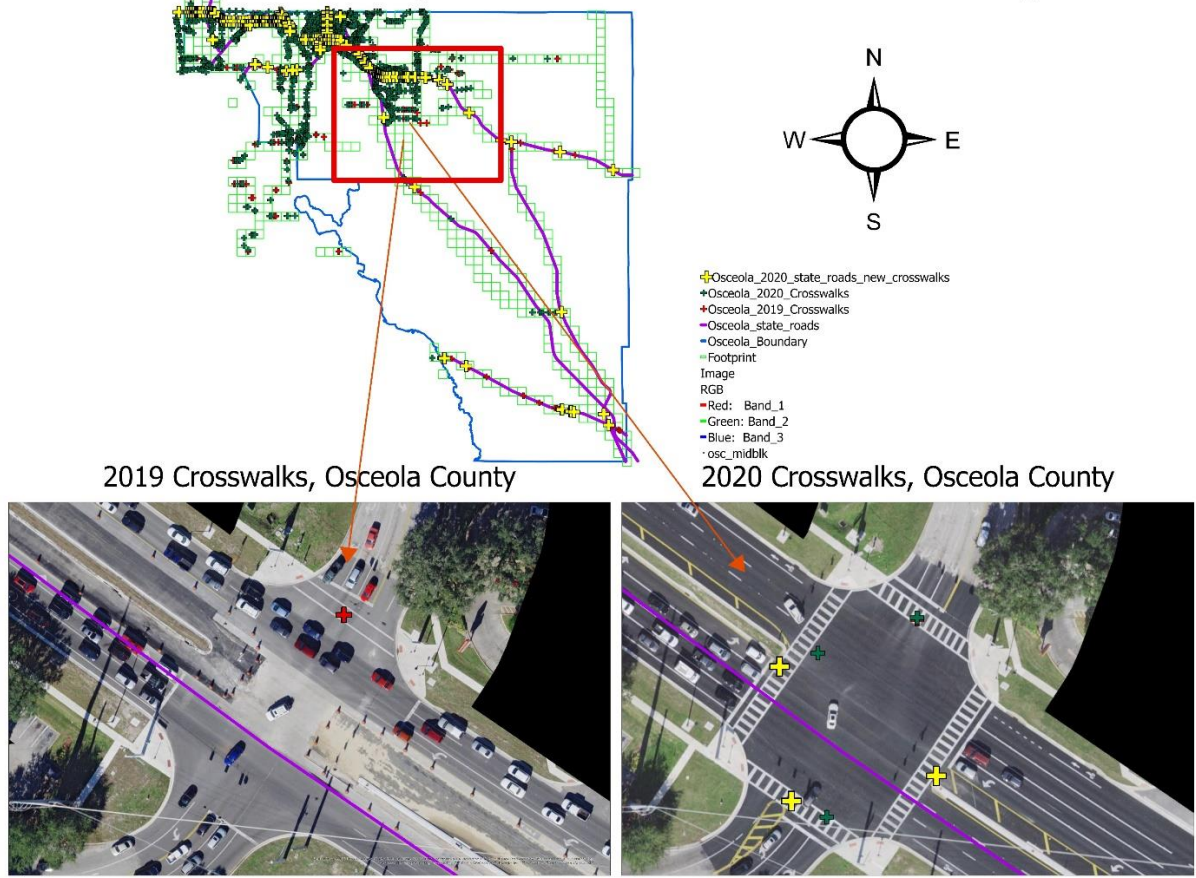

**Figure 9:** 2019 to 2020 Crosswalk Changes in Osceola County

## CONCLUSIONS AND FUTURE WORK

This study explores the application of computer vision tools to detect and extract changes in roadway geometry features, with a specific focus on crosswalks in Florida counties as a proof of concept. This innovative approach utilizes computer vision technology to enable frequent updates to the existing crosswalk inventory, potentially replacing labor- and error-intensive manual methods. In this study, a GIS-based change detection framework was developed to extract crosswalk changes observed from detections carried out in two images with varying temporal resolution. The developed framework was applied to Orange County in Florida. A manual verification was performed on the changes observed in Orange County. The changes were classified into new crosswalks and improved or modified crosswalks. Following this, the framework was automated by developing a model to extract crosswalk changes. The developed model which successfully detect the changes in crosswalks was applied to Seminole and Osceola Counties in Florida.

Findings indicate the model can successfully detect observed crosswalk changes using the CSV file of the old and new detected crosswalks. The output from the observed changes includes both false positives and true positives and will therefore require the post processing stage to eliminate false detections and duplicates. The framework that was created demonstrates excellent performance in detecting changes, including previously overlooked crosswalks from earlier detections. The resulting data can then be seamlessly integrated with the current crosswalk data, ensuring an up-to-date inventory, and preventing any potential data redundancy or loss during the update of roadway characteristics inventory (RCI) data.

The developed system can successfully identify changes in crosswalks extracted from high resolution bi-temporal images. By eliminating the need for manual inventory procedures and enhancing the quality of highway geometry data through the reduction of errors from manual data entry, the results can lead to cost savings for stakeholders. Transportation agencies stand to benefit significantly from this roadway data extraction since it enables the identification of markings previously overlooked, updating the inventory, planning maintenance activities, monitoring infrastructure development, comparing crosswalk locations with other geometric features like school zones, and analyzing crash patterns near these zones.

Based on the outcomes of this study, several noteworthy constraints and recommendations have been identified for future endeavors. The accuracy of change detection relies on several critical factors such as accurate geometric correction and calibration, the presence and quality of ground reference data, the complexity of the landscape and environment, the choice of methods or algorithms utilized, and the expertise and experience of the analyst, and time and cost limitations. Change detection methods and processes can introduce errors, including classification and data extraction inaccuracies. Errors stemming from field surveys, such as the accuracy of ground references, and errors arising from post-processing are also noteworthy (*37*). A notable limitation of the developed framework is its inability to distinguish and eliminate false detections. Accuracy assessment approaches for bi-temporal change detection method include field survey, visual interpretation, and using high resolution images. These approaches can be employed to reduce presence of errors in change detection results. Therefore, it is advisable to manually verify observed changes to mitigate or eliminate the occurrence of false positives. Moreover, employing intelligent change detection methods can enhance data detection and minimize information loss during data processing.

In future research, efforts will be made to enhance and expand this model's capabilities to detect and extract changes in other roadway geometric features. In addition, a similar image processing and data processing approach will be employed to mask the new images and identify crosswalk changes in other counties across Florida. This automated process will be utilized to detect crosswalk changes in these counties, facilitating the comprehensive updating of the existing inventory. Furthermore, future work will involve the integration of the extracted crosswalk points with crash data, traffic data, and data on demographics to conduct more detailed and in-depth analyses. This integration will enable a comprehensive understanding of the crosswalk-related implications for road safety and traffic management.

## ACKNOWLEDGEMENTS

This study was sponsored by the State of Florida Department of Transportation (DOT) grant BED30-962-01. The contents of this paper and discussion represent the authors' opinions and do not reflect the official views of the Florida Department of Transportation.

## AUTHOR CONTRIBUTIONS
The following authors confirm contribution to the paper with regards to Study conception and design: Richard B. Antwi, Samuel Takyi, Alican Karaer, Eren Erman Ozguven, Michael Kimollo, Ren Moses, Maxim A. Dulebenets, and Thobias Sando; Data collection: Richard B. Antwi, Samuel Takyi, Eren Erman Ozguven, Ren Moses, Maxim A. Dulebenets, and Thobias Sando; Analysis and interpretation of results;